\RequirePackage{fix-cm}
\documentclass[smallextended]{svjour3}       
\smartqed  
\usepackage{graphicx}
\usepackage{mathptmx}      

\usepackage{amsfonts}
\usepackage{algorithmic}
\usepackage{lscape}
\usepackage{amsmath}
\usepackage{hyperref}
\numberwithin{equation}{section}
\begin{document}
\title{A Semidefinite Optimization-based Branch-and-Bound Algorithm for Several Reactive Optimal Power Flow Problems
}


\author{Julie Sliwak         \and
        Miguel F. Anjos \and
        Lucas Létocart \and
        Emiliano Traversi
}


\institute{J. Sliwak \at
              RTE (Réseau de Transport d'Electricité), France
           \and
           M.F. Anjos \at
             University of Edinburgh, United Kingdom
            \and
            M.F. Anjos \and J. Sliwak \at
            Polytechnique Montréal, Canada
        \and    
            L. Létocart \and J. Sliwak \and E. Traversi \at
            LIPN, UMR CNRS 7030, Université Sorbonne Paris Nord, France
}

\date{Received: date / Accepted: date}

\titlerunning{A Semidefinite B\&B Algorithm for Several Reactive Optimal Power Flow Problems}

\maketitle

\begin{abstract}
The Reactive Optimal Power Flow (ROPF) problem consists in computing an optimal power generation dispatch for an alternating current transmission network that respects power flow equations and operational constraints. Some means of action on the voltage are modelled in the ROPF problem such as the possible activation of shunts, which implies discrete variables. The ROPF problem belongs to the class of nonconvex MINLPs (Mixed-Integer Nonlinear Problems), which are NP-hard problems. In this paper, we solve three new variants of the ROPF problem by using a semidefinite optimization-based Branch-and-Bound algorithm. We present results on MATPOWER instances and we show that this method can solve to global optimality most instances. On the instances not solved to optimality, our algorithm is able to find solutions with a value better than the ones obtained by a rounding algorithm. We also demonstrate that applying an appropriate clique merging algorithm can significantly speed up the resolution of semidefinite relaxations of ROPF large instances.
\keywords{Mixed-integer nonlinear programming \and Nonconvex programming \and Optimal Power Flow \and Semidefinite programming}
\end{abstract}

\section{Introduction}
\label{intro}
The Optimal Power Flow (OPF) problem, first introduced in \cite{carpentier}, allows to simulate the operation of the power transmission system, both in the long-term  for network development studies and in the short-term for operational safety studies. However, this model can be improved to get closer to the actual functioning of the network. Indeed, there are several means of actions on the network that the OPF problem does not take into account, e.g. activating or desactivating shunts, adjusting transformer ratios, making topological changes. Integrating these means is possible at the cost of introducing discrete variables. The Reactive Optimal Power Flow (ROPF) problem usually focus on the two first means of action: activation of shunts and adjustment of transformer ratios.  
The ROPF is a nonconvex MINLP (Mixed-Integer Nonlinear Programming Problem). Nonconvex MINLPs are NP-hard problems and they are very difficult to solve to global optimality in practice. In some cases, even finding a feasible solution can be a challenge. That is why many research works on ROPF problems only focus on finding a feasible solution thanks to stochastic heuristics \cite{zhou2014strength,lenin2016hybrid,rajan2016exchange,mehdinejad2016solution,heidari2017gaussian,ben2018whale,kilicc2014optimizing,chen2017optimal} or artificial intelligence methods \cite{xu2019optimal}. Other works use a convex relaxation both to get a lower bound and to compute a feasible solution (with round-off procedures): \cite{binganeORPD} proposes a SDP relaxation, \cite{kayacik2020misocp} a SOCP (Second-Order Cone Optimization) relaxation and \cite{hijazi2017convex} a quadratic relaxation which is not strictly speaking applied to the ROPF problem but which could be applied. 
Round-off techniques are proposed in  \cite{capitanescu2010sensitivity} and \cite{yang2017optimal}. Recently, \cite{davoodi2019novel} uses a ROPF model with continuous variables only and proposes to reformulate it with an SDP problem. Another recent work \cite{lopez2021optimal} proposes to use a mixed-integer SDP relaxation and a Branch-and-Bound algorithm to solve it. In this work, a discrete modelling of shunt elements and of tap position of on-load tap changing transformers is used. Feasible solutions are computed using MINLP solvers and do not use the information of the mixed-integer SDP relaxation. This resolution method is tested on networks up to 300 buses. As far as we know, the only heuristic methods present in the literature to solve the ROPF problem by solving a continuous convex relaxation are based on rounding of the obtained solutions. Nevertheless, other methods like spatial Branch-and-Bound algorithms could be applied to the ROPF problem. See \cite{survey_MINLP_nonconvex} for a survey on different methods for generic nonconvex MINLPs.

In this paper, we propose to solve three new variants of the ROPF problem, two of them aiming at studying the impact of the shunt elements to an OPF model and the other focusing on generation. The first variant imposes a constraint on the number of shunts that can be activated. The second one constrains the number of shunts that can be switched from an initial state. The third one imposes a constraint on the active generation to simulate the primary frequency control. 
These three versions of the ROPF problem meet several needs of TSOs (Transmission System Operators) like RTE (Réseau de Transport d'Electricité), the French TSO. Indeed, shunt elements can be quickly damaged if they are constantly solicited. It is therefore preferable for a TSO to limit the activation or the movement of these elements. In case of a voltage problem, it is also important to be able to take into account the constraints on the active power that depend on the planned generation plan. We propose a Branch-and-Bound (B\&B) algorithm based on a SDP relaxation to solve these three nonconvex MINLPs. We test our approach on MATPOWER \cite{matpower} instances and we compare our results to a simple rounding technique from the SDP relaxation solution. 

There are four main contributions in this paper:

1. We introduce practical aspects in the ROPF problem that we model as new constraints: limiting the number of shunt elements that can be activated, limiting the number of shunt elements that can be switched from an initial state, constraining the active generation to simulate the primary frequency control;

2. We propose a flexible SDP-based B\&B method to try to reach global optimality. To the best of our knowledge, it is the first time that a B\&B approach is tested for the ROPF problem. We show that our method allows to close the gap or at least improve it on most instances;

3. We show that the SDP-based B\&B method is able to find heuristic solutions that are better of the ones obtained with rounding heuristics; 

4. We tackle large instances (from six thousands buses until more than ten thousands buses) whereas, as far as we know, the largest instance considered in the literature has a little more than 3000 buses (in \cite{binganeORPD}). We also show that by applying an appropriate clique merging algorithm, we can significantly speed up the resolution of semidefinite relaxations of large ROPF instances.

The rest of this paper is organized as follows. Section 2 describes the three ROPF formulations. The semidefinite relaxation used for these problems is presented in section 3. Section 4 details our B\&B strategy. Computational tests are presented in section 5 and section 6 concludes the paper.

\section{Three variants of the ROPF problem}
\label{sec:ROPF}

In this section, we present the three variants of the ROPF problem that we want to solve. We first describe all the notations used in this paper and then we detail the modelling.

\subsection{Nomenclature}
\label{sec:nomenclature}
\begin{itemize}
    \item $T(N,B)$ is a $n$-buses transmission network. $N$ is the set of buses (nodes) and $B$ is the set of branches (arcs between the buses);
    \item $G \subset N$ represents the subset of generator buses;
    \item $S \subset N$ represents the subset of buses equipped with a shunt element;
    \item At bus $n\in N$, $v_n$ is the complex voltage, $S_n$ the complex power generation and $u_n$ the activation variable of a shunt. $u_n^0$ represents the initial state of a shunt: $u_n^0=1$ if the shunt is activated, $u_n^0=0$ otherwise. $S^{l}_n$ is the complex load and $v_n^{min}$, $v_n^{max}$ are bounds on the voltage magnitude;
    \item At bus $n\in S$, $g_n$ and $b_n$ are respectively shunt conductance and susceptance;
    \item $Re(z)$ (resp. $Im(z)$) denotes the real (resp. imaginary) part of the complex number $z$;
    \item $\mathbf{j}$ denotes the complex number such that $\mathbf{j}^2=-1$;
    \item $\overline{z}$ represents the conjugate of the complex number $z$ and $|z|$ its modulus;
    \item $Z^H$ denotes the transpose conjugate of the complex matrix $Z$;
    \item For a generator $g \in G$, $c_g$ and $k_g$ represent the linear and the constant cost and $P_g^{min}$, $P_g^{max}$  (resp. $Q_g^{min}$, $Q_g^{max}$) the bounds on the active (resp. reactive) power. $P_g^0$ represents the planned active generation;
    \item For a branch $l \in B$, $o_l$ is the origin of branch $l$ and $d_l$ its destination. $y_l$ stands for the admittance, $b_l$ the charging susceptance, $\tau_l$ the transformer ratio, $\theta_l$ the transformer phase shift angle and $i_l^{max}$ for the current magnitude limit for the branch.
\end{itemize}

\subsection{Modelling}
\label{sec:modelization}

Power transmission networks can be modelled as oriented graphs $T=(N,B)$. A node $n \in N$ is called a bus and represents an electrical node of the power transmission network. A bus can be a generator bus if a generation mean is associated to it (power plant, solar pannels, etc). To simplify, we suppose that a bus can only have one generation mean (if several ones are given for the same bus in the MATPOWER input, we aggregate them using the cost of the last one in the MATPOWER list and we use the sums of power limits as power bounds). The set $G \subset N$ denotes the set of all generator buses. Some buses must also meet a power demand, which is called a load. Finally, some buses are equipped with shunt elements which are electrical components that indirectly allow to control the voltage. The set $S \subset N$ denotes the set of all buses equipped with a shunt element. An arc $b\in B$ is called a branch and represents a transmission line and/or a transformer. Power transmission networks using Alternating Current (AC) transmission lines, all physical quantities as voltage, current and power are complex numbers. The goal of the ROPF problem is to compute a network state that satisfies power flow equations and safety constraints while minimizing generation costs. To define such a state, we need to know: \begin{itemize}
    \item the complex voltage $v_n$ at each bus $n\in N$;
    \item the complex power $S_g$ generated  at each generator bus $g\in G$;
    \item the binary state (on/off) $u_s$ of each shunt element $s\in S$.
\end{itemize} These three quantities are the variables of the ROPF problem \ref{ROPF}. Other physical parameters of the power network such as the current on branches can be deduced from these. 

The first constraints that these three variables have to satisfy are the power flow equations: the power going into a bus must be equal to the power coming out. More precisely, the power $S_n$ generated at a bus $n \in N$ is equal to the load $S_n^l$ plus the power induced by shunt elements $(g_n-\mathbf{j}b_n)|v_n|^2  u_n$ plus the power induced by outgoing and incoming branches. The power at bus $n$ for an outgoing branch~$l=(n,d)$ depends on physical parameters of the branch~$l$ (the admittance $y_l$ , the charging susceptance $b_l$, the transformer ratio $\tau_l$ and the transformer phase shift angle $\theta_l$) and quadratically depends on the voltage in the following way:  $S^{orig}_l(v_n,v_d) = \frac{\overline{y_l} - \mathbf{j}b_l}{\tau_l^2}{|v_n|^2}-\overline{y_l} \frac{e^{\mathbf{j}\theta_l}}{\tau_l }v_n\overline{v_d}$. The power at bus $n$ for an incoming branch $l=(o,n)$ depends on the same physical parameters and also quadratically depends on the voltage:
$S^{dest}_l(v_o, v_n) = -\overline{y_l} \frac{e^{-\mathbf{j}\theta_l}}{\tau_l }\overline{v_o}v_n+(\overline{y_l} - \mathbf{j}b_l) |v_n|^2$.

Next there are different types of safety constraints. Firstly, there are constraints on the power generation which correspond to real limits of generation groups. If the bus $n$ is a generator bus, i.e., $n \in G$, there are bounds on the generation: the real part of $S_n$ must be between $P_n^{min}$ and $P_n^{max}$ whereas the imaginary part of $S_n$ must be between $Q_n^{min}$ and $Q_n^{max}$. Logically, if the bus $n$ is not a generator bus, i.e. $n \notin G$ then it cannot generate any power so $S_n=0$. Secondly, there are  bounds $v_n^{min}$, $v_n^{max}$ on the voltage magnitude which are operating standards to prevent damage to equipment or power outages. In the optimization model, these bounds constraints are written with the squared voltage magnitude to avoid squared roots. Thirdly, there are bounds $i_l^{max}$ on the current magnitude on branches which model thermal limits. In the literature, these thermal limits can also be modelled in term of power but the limits in current reflect the practice at RTE. The incoming and outgoing currents on a branch $l$ depend on the same physical parameters as for the power $S^{orig}_l$ and $S^{dest}_l$ since the power is equal to the voltage times the conjugate of the current. More precisely, the outgoing current for branch $l=(o_l, d_l)$ is $i^{orig}_l = \frac{\overline{y_l} + \mathbf{j}b_l}{\tau_l^2}{v_{o_l}}-\overline{y_l} \frac{ e^{-\mathbf{j}\theta_l}}{\tau_l}v_{d_l}$ and the incoming current from branch $l$ is $i^{dest}_l=-\overline{y_l} \frac{e^{\mathbf{j}\theta_l}}{\tau_l }v_{o_l}+(\overline{y_l} + \mathbf{j}b_l)v_{d_l}$. Once again, we model these constraints with the squared of the current magnitude to avoid squared roots.

Finally, we add three optional constraints. The first one, described in Equation \eqref{MAXkshunts}, is a constraint regarding shunt elements. In the literature, all ROPF problems allow to activate as many shunt elements as one wants, as if there were no cost. In practice, activating shunt elements has a non-financial cost and TSOs like RTE are interested in activating as little as possible to avoid premature aging of the material while computing a satisfying generation dispatch. To tackle this issue, we add a sum constraint to activate at most $k$ shunt elements: 
\begin{equation}
\sum\limits_{n \in S} u_n \leq k.
\tag{MAXkshunts}
\label{MAXkshunts}
\end{equation}
The parameter $k$ is chosen arbitrarily. In our studies, we take $k=4$ because it is a value used at RTE to limit the switchs of shunts over a day in long-term simulations. If $k=4$ leads to infeasibility (proved or supposed), we define $k$ as the smallest value for which a feasible solution can be computed.

The second constraint, described in Equation \eqref{MAXkmoves}, also concerns shunt elements but addresses a shorter-term problem. Unlike the previous one, this constraint takes into account the initial configuration for the shunt elements: some shunts are activated and the others are not. This initial configuration is described by the vector $u^0$. The constraint consists in limiting the number of binary variables flipping their value (with respect to $u^0$) either from 1 to 0 or from 0 to 1:
\begin{equation}
\sum\limits_{n \in S:u_n^0=0} u_n + \sum\limits_{n \in S:u_n^0=1} (1-u_n) \leq k.
\tag{MAXkmoves}
\label{MAXkmoves}
\end{equation}
This type of constraint is usually called local branching constraint in mixed-integer linear programming \cite{fischetti2003local}. The parameter $k$ is chosen arbitrarily, as for constraint \eqref{MAXkshunts}. We take $k=4$ in our tests since it reflects the practice at RTE. 

The third type of optional constraint, described in Equation \eqref{GENmoves}, concerns the active power variables and approximates primary frequency control by allowing a uniform ascent or descent of the  active generation plan defined by the vector $P^0$. More precisely, two binary variables $\delta^+$ and $\delta^-$ indicate if the generation plan is increased or decreased. These two variables are opposed: if $\delta^+ = 1$ then $\delta^- = 0$ and vice versa. This is enforced by a constraint on the sum of both. This models the fact that the generators all go in the same direction to compensate for a lack or excess of generation. In addition, they must all contribute in a "uniform" manner according to their power bounds. We model this by an affine constraint depending on a unique variable $\lambda^+$ or $\lambda^-$ which makes all the generators reach their upper (or lower) bound at the same time ($\lambda^+ =1$ or $\lambda^- =0$). The variables $\lambda^+$ and $\lambda^-$ have bounds such as the active power $Re(S_n)$ is always between its bounds $P_n^{min}$ and $P_n^{max}$. Therefore, if the constraints in \eqref{GENmoves} are used, the constraints $P_n^{min} \leq Re(S_n)\leq P_n^{max}$ are redundant. To summarize, the constraints \eqref{GENmoves} make use of four additional variables including two binary ones, and constrain the active generation with respect to a given generation plan. These constraints are detailed below:

\begin{equation}
\left\{
\begin{array}{ll}
Re(S_n)= [P_n^{min}+2(P_n^0-P_n^{min})\lambda^-]\delta^-\\
+ [2P_n^0-P_n^{max}+2(P_n^{max}-P_n^0)\lambda^+]\delta^+ &  \forall n \in G\\
 0 \leq \lambda^- \leq 0.5\\
 0.5 \leq \lambda^+ \leq 1\\
 \delta^+ + \delta^- = 1\\
 \lambda^-, \lambda^+ \in \mathbb{R}\\
\delta^+ , \delta^-  \in \{0,1\}.\\
\end{array}
\right.
\tag{GENmoves}
\label{GENmoves}
\end{equation}

These three optional new sets of constraints are one of the two major changes compared to the ROPF formulations in the literature. The other major change is that we do not make the transformer tap ratios variable because it is more useful for RTE to have reliable results with one main means of voltage control rather than uncertain results with several means. 

The problem objective is to minimize total generation cost, that is, a function depending on the real part of the powers $S_g$ for generator buses $g\in G$. In the literature, this function is usually quadratic. However, we use linear generation costs $c_g, k_g$ to get closer to the practice at RTE which usually consists in minimizing the standard deviation from a given generation dispatch.

To summarize, the ROPF problem we tackle is defined as follows for a transmission network $T(N,B)$:  

\begin{equation}
\begin{array}{lll}
\min\limits_{v, S,u} & \sum\limits_{g\in G}c_g(Re(S_g))+k_g& \\[4mm]
s.c. & S_n = S_n^{l}+(g_n-\mathbf{j}b_n)|v_n|^2  u_n  +\sum\limits_{l=(n,d)}(\frac{\overline{y_l} - \mathbf{j}b_l}{\tau_l^2}{|v_n|^2}-\overline{y_l} \frac{e^{\mathbf{j}\theta_l}}{\tau_l }v_n\overline{v_d}) \\
& -\sum\limits_{l=(o,n)}(-\overline{y_l} \frac{e^{-\mathbf{j}\theta_l}}{\tau_l }\overline{v_o}v_n+(\overline{y_l} - \mathbf{j}b_l) |v_n|^2) & \forall n \in N\\[4mm]
& P_n^{min} \leq Re(S_n)\leq P_n^{max} & \forall n \in G\\[2mm]
& Q_n^{min} \leq Im(S_n)\leq Q_n^{max} & \forall n \in G\\[2mm]
& S_n=0 &  \forall n \notin G\\[2mm]
& (v_n^{min})^2 \leq |v_n|^2 \leq (v_n^{max})^2& \forall n \in N \\[2mm]
& |\frac{\overline{y_l} + \mathbf{j}b_l}{\tau_l^2}{v_{o_l}}-\overline{y_l} \frac{ e^{-\mathbf{j}\theta_l}}{\tau_l}v_{d_l}|^2 \leq (i_l^{max})^2 & \forall l \in B\\[2mm]
& |-\overline{y_l} \frac{e^{\mathbf{j}\theta_l}}{\tau_l }v_{o_l}+(\overline{y_l} + \mathbf{j}b_l)v_{d_l}|^2 \leq (i_l^{max})^2 & \forall l \in B\\[2mm]
& \eqref{MAXkshunts} \textsf{ OR } \eqref{MAXkmoves} \textsf{ OR } \eqref{GENmoves}\\[2mm] 
& u_n=0 &  \forall n \notin S\\[2mm]
& v_n \in \mathbb{C}, S_n \in \mathbb{C} & \forall n \in N\\[2mm]
&  u_n \in \{0,1\} & \forall n \in S.
\end{array}
\label{ROPF}
\end{equation}

In this formulation, the power generation variables can be eliminated. The costs being linear, the resulting problem is a nonconvex Mixed-Integer Quadratically Constrained Quadratic Problem (MIQCQP) with binary and complex variables.

\section{A semidefinite relaxation for the ROPF problem}
\label{sec:SDP}
In this section, we present a SDP relaxation for the ROPF problem described in \ref{ROPF}. This SDP relaxation is used in our B\&B procedure to compute lower bounds.

To construct the SDP relaxation of problem \ref{ROPF}, the Hermitian matrix $V=vv^H$ is introduced and is relaxed into $V \succeq 0 $ to handle continuous variables. As for binary variables, the variables $u_s$ are relaxed for each shunt element $s\in S$ but new variables $\xi_s$ have to be introduced to replace the quadratic term $u_sV_{ss}$. The quadratic constraints $\xi_s = u_sV_{ss}$ are relaxed using McCormick envelopes \cite{mccormick}. For a product $w=xy$ with $x \in [\underline{x}, \overline{x}]$ and $y \in [\underline{y}, \overline{y}]$, the McCormick envelopes are detailed in Equation \ref{McC}.

\begin{equation}
w = xy \Rightarrow
 \left\{ \begin{array}{l}
w \leq \overline{x}y+x\underline{y}-\overline{x}\underline{y}\\
w \leq x\overline{y}+\underline{x}y-\underline{x}\overline{y}\\
w \geq \underline{x}y+x\underline{y}-\underline{x}\underline{y}\\
w \geq \overline{x}y+x\overline{y}-\overline{x}\overline{y}
\end{array}\right.
\label{McC}
\end{equation}

For each shunt element $s \in S$, $u_s \in [0,1]$ and $V_{ss} \in [(v_s^{min})^2,(v_s^{max})^2]$, then the McCormick envelopes for $\xi_s = u_sV_{ss}$ are:

\begin{equation}
     \left\{ \begin{array}{l}
\xi_s \leq V_{ss}+(v_s^{min})^2(u_s-1)\\
\xi_s \leq (v_s^{max})^2 u_s\\
\xi_s \geq (v_s^{max})^2 (u_s-1) + V_{ss}\\
\xi_s \geq (v_s^{min})^2u_s
\end{array}\right.
.
\label{McC$_s$}
\end{equation}

The SDP relaxation obtained is as follows:
\begin{equation}
\begin{array}{lll}
\min\limits_{V, S,u, \xi} & \sum\limits_{g\in G}c_g(Re(S_g))+k_g& \\[4mm]
s.c. &  S_n = S_n^{l} + (g_n-\mathbf{j}b_n)\xi_n + \sum\limits_{l=(n,d)}(\frac{\overline{y_l} - \mathbf{j}b_l}{\tau_l^2}{V_{nn}}-\overline{y_l} \frac{e^{\mathbf{j}\theta_l}}{\tau_l }V_{nd})\\
& - \sum\limits_{l=(o,n)}(-\overline{y_l} \frac{e^{-\mathbf{j}\theta_l}}{\tau_l }V_{no}+(\overline{y_l} - \mathbf{j}b_l) V_{nn}) &\forall n \in N\\[4mm]
& P_n^{min} \leq Re(S_n)\leq P_n^{max} & \forall n \in G\\[2mm]
& Q_n^{min} \leq Im(S_n)\leq Q_n^{max} & \forall n \in G\\[2mm]
& (v_n^{min})^2 \leq V_{nn} \leq (v_n^{max})^2& \forall n \in N \\[2mm]
& I^{orig}_l \cdot V \leq (i_l^{max})^2 & \forall l \in B\\[2mm]
& I^{dest}_l \cdot V \leq (i_l^{max})^2 & \forall l \in B\\[2mm]
& S_n=0 &  \forall n \notin G\\[2mm]
& \eqref{MAXkshunts} \textsf{ OR } \eqref{MAXkmoves}  \textsf{ OR } \eqref{GENmoves}\\[2mm] 
&      \left\{ \begin{array}{l}
\xi_s \leq V_{ss}+(v_s^{min})^2(u_s-1)\\
\xi_s \leq (v_s^{max})^2 u_s\\
\xi_s \geq (v_s^{max})^2 (u_s-1) + V_{ss}\\
\xi_s \geq (v_s^{min})^2u_s
\end{array}\right. & \forall s \in S\\
& u_n=0 &  \forall n \notin S\\[2mm]
&  \xi_n = 0 &  \forall n \notin S\\[2mm]
& V \succeq 0 \\[2mm]
& S_n \in \mathbb{C} & \forall n \in N\\[2mm]
&  u_n \in [0,1] & \forall n \in S
\end{array}
\label{SDP}
\end{equation}
where for each branch $l\in B$, $I^{orig}_l$ (resp. $I^{dest}_l)$ is the Hermitian matrix such that  \newline $I^{orig}_l \cdot vv^H= |\frac{\overline{y_l} + \mathbf{j}b_l}{\tau_l^2}{v_{o_l}}-\overline{y_l} \frac{ e^{-\mathbf{j}\theta_l}}{\tau_l}v_{d_l}|^2$ (resp. $I^{dest}_l \cdot vv^H = |-\overline{y_l} \frac{e^{\mathbf{j}\theta_l}}{\tau_l }v_{o_l}+(\overline{y_l} + \mathbf{j}b_l)v_{d_l}|^2$).

The proposed formulation provides strong lower bounds. On the other hand, it is not easy for the given model to further strengthen the provided model. More precisely, we tested different techniques to tighten this SDP relaxation. We tried to tighten the bounds on $V_{ss}$ variables with Optimality-Based Bound Tightening (OBBT) techniques \cite{OBBT} in order to tighten the McCormick envelopes. We also quadratized the constraint \eqref{MAXkmoves} replacing this linear constraint by $2|S|$ quadratic constraints modelled with an SDP variable. However, none of these techniques improved significantly the lower bound given by the SDP relaxation.

Finally, this complex SDP relaxation can be converted to a SDP problem with real variables using the rectangular formulation of a complex number $z=Re(z)+\mathbf{j}Im(z)$ and a matrix of size $2n$. Subsequently, it can be reformulated using clique decomposition techniques \cite{fukudaI,fukudaII,chordal_graphs_SDP} to speed up the resolution. 

\section{An SDP-based Branch-and-Bound algorithm}
\label{sec:BandB}
In this section, we detail our flexible strategy to solve all variants of the ROPF problem detailed in \ref{ROPF}. This strategy is made in two steps: we first compute a pair of lower and upper bounds of the optimal objective, then if the gap between the two bounds is not satisfying, we apply a SDP-based B\&B algorithm on a restricted number of shunt elements in order to improve or close the gap.

To begin, we compute a valid upper bound with the following three-steps heuristic based on the nonlinear local solver KNITRO \cite{KNITRO}:
\begin{itemize}
    \item 
In a first step, the continuous relaxation of problem \ref{ROPF} is solved to find a relevant starting point for the resolution of the MINLP. 
\item In a second step, the MINLP is solved using a MPEC (Mathematical Programming with Equilibrium Constraints) reformulation: the binary variables are replaced by complementary constraints ($x(x-1)=0$) which are penalized into the objective function and the penalization term is updated throughout the resolution. This update is done automatically by KNITRO.
\item In a third step, the binary variables are fixed by rounding the solution of the second step and the resulting continuous problem is solved. This last step ensures feasibility.
\end{itemize}

We then compute a lower bound by solving the SDP relaxation detailed in \ref{SDP} using classical clique decomposition techniques. To accelerate the resolution for instances with more than 1000 buses, we use a tailored clique merging algorithm presented in \cite{sliwak2020clique}. 

If the gap between the three-steps heuristic and the SDP relaxation is not satisfying, we apply the generic B\&B algorithm presented in Figure \ref{BB}. This algorithm takes as input an instance, the solution of the SDP relaxation detailed in \ref{SDP} and two real thresholds $l$ and $u$, both between 0 and 1. It returns an upper bound of the optimal objective. The first step in the algorithm (line 1) consists in fixing the binary variables whose value in the SDP relaxation is greater than the threshold $u$ and less than the threshold $l$. This sets a first dictionary of fixed variables (line 2). A first upper bound is computed from this dictionary of fixed variables (line 3). The function \texttt{solve\_MINLP} applies our three-steps heuristic if some binary variables are free and only the third step if all binary variables are fixed. The following line (line 4) initializes the B\&B algorithm. We use a node structure that is defined by two arguments: the lower bound of the father's node and the dictionary of fixed variables. We define \texttt{nodes\_list} as the list of nodes that have to be explored. The first node in the list is defined by a father's lower bound set to $-\infty$ and the first dictionary of fixed variables. Until there are some nodes to explore, that is, as long as the list of nodes is not empty, the procedure is the following. First, a node is chosen following a deep first strategy by preferring the node that has more variables fixed to 1 (line 6). The dictionary of fixed variables of this unique node is used for the following computations (line 7) and this node is deleted (line 8). The function \texttt{solve\_SDP} computes a lower bound of the optimal objective using the SDP relaxation detailed in \ref{SDP} and according to the dictionary of fixed variables. There are three cases:
\begin{itemize}
    \item If the lower bound is strictly greater than the current upper bound (NB: this condition can be relaxed to the targeted precision) or if the SDP relaxation is infeasible, the node is pruned, that is, we explore no further (lines 10:11).
    \item If the variables $u_n$ are binary (within a given accuracy) in the SDP relaxation, an upper bound is computed using our function \texttt{solve\_MINLP} and according to the dictionary of fixed variables. If this computation improves the upper bound, we update the current upper bound and we check if some nodes can be pruned because their father's lower bound is strictly greater than the current upper bound (lines 12:16).
    \item Otherwise, two nodes are created by branching on a given binary variable $u_i$. Both have their father's lower bound equal to the lower bound computed using the SDP relaxation but their dictionary of fixed variables differs by one value: it is equal to the current dictionary with one more entry, one corresponding to $u_i=1$ and the other corresponding to $u_i=0$. We  branch on the variable $u_i$ which is the closer to 1 in the SDP solution (lines 17:24).   
\end{itemize}

\renewcommand{\algorithmicrequire}{\textbf{Input:}}
\renewcommand{\algorithmicensure}{\textbf{Output:}}
\begin{figure}
\centering
\begin{algorithmic}[1]
\REQUIRE Instance, SDP relaxation solution ($\xi_n^* / V_{nn}^*$), a lower threshold $l$, an upper threshold $u$
\ENSURE UB
\STATE Fix variables $u_n$ to 0 if $\xi_n^* / V_{nn}^* \leq l$ and to 1 if  $\xi_n^* / V_{nn}^* \geq u$
\STATE fixing0 $\leftarrow$ dictionary of all fixed variables
\STATE UB $\leftarrow$ \texttt{solve\_MINLP}(instance, fixing0)
\STATE nodes\_list $\leftarrow$ [node(-$\infty$, fixing0)]
\WHILE{$|$nodes\_list$| > 0$ }
\STATE Select node n $\in$ nodes\_list following deep first strategy
\STATE fixing $\leftarrow$ dictionary of all fixed variables at node n
\STATE Delete node n
\STATE LB = \texttt{solve\_SDP}(instance, fixing)
\IF{LB$>$UB or SDP not feasible} 
\STATE	CUT
\ELSIF{variables $u_n$ are binary in the SDP solution}
\STATE UB' $\leftarrow$ \texttt{solve\_MINLP}(instance, fixing)
\STATE UB $\leftarrow$ UB' if UB'$<$ UB
\STATE Delete nodes  in nodes\_list whose $LB$ is strictly greater than $UB$
\STATE CUT
\ELSE 
\STATE CREATE TWO NODES
\STATE Branch on variable $u_i$ which is the closer to 1 in the SDP solution
\STATE Create two copies fixing1 and fixing0 of fixing
\STATE Fix $u_i$ to 1 in fixing1
\STATE Fix $u_i$ to 0 in fixing0
\STATE Add node(LB, fixing1) to nodes\_list
\STATE Add node(LB, fixing0) to nodes\_list
\ENDIF
\ENDWHILE
\end{algorithmic}
\caption{Pseudo-code of our SDP-based B\&B algorithm}
\label{BB}
\end{figure}

Our B\&B algorithm uses the SDP relaxation presented in \ref{SDP} to compute a lower bound at each node. An SDP relaxation is much more costly to solve than a linear relaxation but it provides tighter lower bounds and it has an interest to use this relaxation only if the exploration tree has a reasonable size. That is why we have not implemented a spatial B\&B and we only branch on the binary variables. Therefore, the presented method is not exact: even if all binary variables are fixed, there may still be an optimality gap between the upper bound and the lower bound - if the SDP relaxation is not exact or if the feasible solution computed with a nonlinear local solver is not a global optimum. For this reason, our objective here is to find the best upper bound, guided by the SDP relaxation, in a reasonable amount of time. Indeed, TSOs like RTE do not seek to obtain the global optimal solution at all costs and they usually prefer to have good solutions fast enough. Since a B\&B algorithm is costly, especially when the number of binary variables increases, we propose to use the SDP relaxation solution to fix a certain number of binary variables in order to reduce the exploration tree. Indeed, without fixing any binary variables, we take the risk of having very long procedures and potentially no improvement at the end: we have no guarantee that we have found the optimal solution at the end of the B\&B procedure since the problem \ref{ROPF} is nonconvex, which implies that the relaxation at a leaf node may not be exact or that the solution computed by the local solver may not be a global optimum. 
Therefore, we decide to avoid the use of a 
spatial B\&B 
and we opt for a matheuristic approach, able to provide a good
trade-off between computation time and quality of the solution. 

Finally, we propose to adapt the fixing strategy (line 1 of Algorithm~\ref{BB}) to the type of ROPF problem as follows:
\begin{itemize}
    \item 
For the ROPF with \eqref{MAXkshunts}, the constraint implies that many binary variables will be 0 so we propose to fix to 0 binary variables whose value in the SDP solution is less or equal to 0.25. We choose 0.25 as a threshold because it reduces significantly the number of binary variables while leaving a bit of choice for the shunt elements to be activated. Thus, our B\&B is supposed to converge fairly quickly but it still has a chance to improve the solution.
\item For the ROPF with \eqref{MAXkmoves}, we compute an initial state by rounding the solution of the continuous relaxation of problem \ref{ROPF} without any optional constraint. This initial state contains lots of activated shunts. That is why we propose to fix to 1 binary variables whose value in the SDP solution is greater or equal to 0.75. This threshold gives a good trade-off between the number of fixed variables and the number of free variables.
\item For the ROPF with \eqref{GENmoves}, there is more freedom on the shunt elements. We therefore opt for a slightly different strategy: we propose to fix to 1 binary variables whose value in the SDP is greater to 0.9 and to 0 binary variables whose value in the SDP is less than $10^{-4}$. We have chosen these thresholds by testing the impact of the fixation on the lower bound. For thresholds 0.9 and $10^{-4}$, the lower bound is not significantly deteriorated.  
\end{itemize}

\section{Computational results}
\label{sec:results}

In this section, we present computational results for the three variants of the ROPF problem. For all problems, the tests were carried out on a Processor Intel® Core™ i7-6820HQ CPU @2.70GHz. We used Julia 1.0.3. \cite{Julia} to implement our B\&B algorithm. We constructed the ROPF instances with our module MathProgComplex.jl \cite{MathProgComplex_powertech}. The packages JuMP.jl 0.19.0 \cite{jump} and Mosek.jl with MOSEK 9.1 \cite{mosek} were used for the SDP resolution. The modelling language AMPL Version 2018062 \cite{AMPL} along with the solver KNITRO 11.0.1 were used for the MINLP local resolution. 

In Section~\ref{sec:clique_merging}  we first show that the SDP resolution time is improved by applying a relevant clique merging algorithm. We then present in Section~\ref{sec:resultsBB} the results of our B\&B algorithm.

\subsection{Impact of clique merging on the SDP computation time}\label{sec:clique_merging}
We compare MOSEK resolution time for two clique decomposition strategies on MATPOWER instances with more than 1000 buses. The first strategy does not use clique merging: maximal cliques are computed from a Cholesky factorization with an AMD (Approximate Minimum Degree) ordering \cite{AMD}. The second strategy consists in applying a clique merging algorithm on the first clique decomposition, which leads to a trade-off between the number and the size of cliques and the number of linking constraints. This clique merging algorithm is described in \cite{sliwak2020clique} and we take $k_{max}=1$ in all our experiments.

Table \ref{SDPtime} shows the results of this comparison for the ROPF problem with optional constraint \eqref{MAXkshunts}. We observe that the resolution time is significantly decreased by the clique merging algorithm. Roughly speaking, it is divided by two. More precisely, the mean improvement is of 129\%. This significant reduction is all the more important as a SDP relaxation has to be solved at each node of a B\&B algorithm. Note that similar results are observed for the other two variants of the problem.

\begin{table}
\caption{MOSEK Computation Time for MATPOWER Instances with More Than 1000 Buses}
\label{SDPtime}
\centering
\begin{tabular}{lr|rr}
\hline\noalign{\smallskip}
Instance & $k$ & \multicolumn{2}{c}{MOSEK resolution time (seconds)}\\
 & & No clique merging & Clique merging\\
\noalign{\smallskip}\hline\noalign{\smallskip}
case1354pegase    & 4    & 12.28  & 6.72   \\
case1888rte       & 4    & 11.50  & 7.09   \\
case1951rte       & 4    & 11.50  & 8.41   \\
case\_ACTIVSg2000 & 4    & 475.76 & 86.45  \\
case2383wp        & 4    & 93.61  & 29.36  \\
case2736sp        & 4    & 105.77 & 49.55  \\
case2737sop       & 4    & 119.11 & 47.34  \\
case2746wop       & 4    & 115.67 & 66.80  \\
case2746wp        & 4    & 96.44  & 53.11  \\
case2848rte       & 4    & 29.58  & 11.69  \\
case2868rte       & 4    & 21.88  & 11.16  \\
case2869pegase    & 12    & 48.17  & 26.05  \\
case3012wp        & 4    & 151.55 & 62.19  \\
case3120sp        & 4    & 167.83 & 73.19  \\
case3375wp        & 4    & 120.36 & 60.52  \\
case6468rte       & 4    & 259.48 & 111.08 \\
case6470rte       & 4    & 243.70 & 113.14 \\
case6495rte       & 14   & 226.61 & 111.77 \\
case6515rte       & 66   & 199.13 & 95.11  \\
case9241pegase    & 125  & 922.20 & 388.76 \\
case13659pegase   & 1100 & 269.56 & 123.63\\
\noalign{\smallskip}\hline
\end{tabular}
\end{table}

\

\subsection{Performance of Algorithm~\ref{BB}}
\label{sec:resultsBB}

We decide to compare the performance of our algorithm with the only alternative present in the literature for this class of problems: a rounding heuristic. This rounding heuristic consists in fixing to 1 all binary variables greater or equal to 0.5 (the $k$ greatest ones if there are more than $k$ for the ROPF problem with constraint \eqref{MAXkshunts}) and the rest to 0 and using a nonlinear local solver (KNITRO) to compute a feasible solution. The main tables of this section 
are Table~\ref{ResultsMAXkshunts},
Table~\ref{ResultsMAXkmoves},
Table~\ref{ResultsGENmove1} and
Table~\ref{ResultsGENmove2}. They all have the same structure. The first columns give information about the MATPOWER instances: the name of the instance, $|S|$ the number of shunt elements and optionally another indication about the variant. The following three columns show the bounding of the optimal objective computed within the first phase of our algorithm, before the B\&B algorithm: the upper bound $UB$ computed by KNITRO with our three-steps heuristic, the lower bound $LB$ computed by MOSEK and the relative optimality gap in percentage given by the formula $\frac{\textsf{UB-LB}}{\textsf{UB}}$. The following five columns show the results of our B\&B algorithm on instances for which the gap computed with the first step of our algorithm is not satisfying. We consider that an instance is solved if the relative gap is less or equal to $10^{-4}$, which is an acceptable error for RTE. The first column of the B\&B table part presents the number of binary variables that are free in our B\&B algorithm. The two following columns display the number of nodes explored in the B\&B tree and the time in seconds that it takes to perform the B\&B. We use a time limit of 3600 seconds but some instances do not end within this time limit; for these instances, we indicate '$>$3600'. The  following column shows the best upper bound computed in our algorithm: in the B\&B algorithm within one hour or within the first step if the upper bound computed in the B\&B algorithm is not as good, which can happen since binary variables are fixed from the very beginning in our B\&B algorithm. The next column presents the relative optimality gap obtained with this new upper bound. The last two columns indicate the results found with a rounding heuristic: the upper bound computed and the corresponding relative optimality gap.

\subsubsection{The ROPF problem with constraint \eqref{MAXkshunts}}
\label{sec:resultsMAXkshunts}

We first apply our method on the ROPF problem with \eqref{MAXkshunts}. As mentioned in section \ref{sec:modelization}, we use $k=4$ by default because it is a value that reflects the practice at RTE. If $k=4$ leads to infeasibility (proved or supposed), we define $k$ as the smallest value for which we can compute a feasible solution. 

Results of this comparison are presented in Table \ref{ResultsMAXkshunts}. We observe that out of 31 instances, 13 are solved within the first phase of our procedure and 7 more thanks to the B\&B algorithm, for a total of 20 solved instances compared to 14 with the rounding procedure. Moreover, our procedure is more robust than a rounding procedure: there are 9 instances for which the rounding procedure has not found a feasible solution versus 1 for our procedure (case6470rte). However, if  the time limit is doubled, a feasible solution is found for the instance case6470rte (UB=98697.64, gap=0.05\%) and it exactly takes 5382 seconds. It is also interesting to see that for large instances (more than 6000 buses), the rounding procedure is often inefficient to compute a feasible solution (one possibility is that the problem is not feasible but we cannot be sure) while our three-steps procedure often finds a feasible solution. The relative gap obtained for the two bigger instances is greater than 1\% but the worst gap for the other ones is 0.52\%. Our algorithm is therefore able to compute good and robustly feasible solutions, which is an important point for RTE. As for computation time for the B\&B algorithm, we observe that instances that are solved to global optimality are solved within 500 seconds. For other instances (except case\_ACTIVSg500), the B\&B is not finished in 3600 seconds. However, if the time limit is sufficiently increased, some B\&B algorithms come to and end: case1888rte (19770 seconds), case\_ACTIVSg2000 (7249 seconds), case2848rte (6747 seconds), case3120sp (11450 seconds) but there is no improvement in the upper bound. For the large instances with more than 6000 buses, the B\&B takes at least 8 hours and the upper bound is not always improved after 8 hours of computations.

\subsubsection{The ROPF problem with constraint \eqref{MAXkmoves}}
\label{sec:resultsMAXkmoves}
We now test our SDP-based B\&B method on the ROPF problem with \eqref{MAXkmoves}. As mentioned in section \ref{sec:modelization}, we take $k=4$ since it is a value that reflects the practice at RTE. We compute an initial state $u^0$ by rounding the solution of the continuous relaxation of the problem \ref{ROPF} without any optional constraint. We could have chosen a random initial state but we preferred to have something more realistic.

Results of this comparison are presented in Table \ref{ResultsMAXkmoves}.  We observe that out of 31 instances, 18 are solved within the first phase of our procedure and 3 more thanks to the B\&B algorithm, for a total of 21 solved instances compared to 16 with the rounding procedure. Besides, our procedure computes a feasible solution for each instance which is not the case for the rounding procedure. The rounding heuristic may even lead to an infeasible problem (case2848rte and case3012wp). Finally, even if the optimality gap is not closed for all instances, our procedure gives reasonable gaps: less than 1\% for instances with less than 9000 buses and 1.5\% at worst for the two largest instances. However, there are 3 instances for which the B\&B procedure did not improve the solution. There are several possible explanations: it can be due to the time limit, it could also reflect the nonconvexity of the problem or it could come from a too restrictive fixation. It is an inherent risk in our inexact B\&B method. This is why we are convinced that a time limit should be imposed to avoid spending a lot of time without improving anything. To summarize, we could say that our procedure is more robust than a rounding procedure and it allows at best to close the gap, at worst to obtain a good feasible solution.

\subsubsection{The ROPF problem with constraints \eqref{GENmoves}}
\label{sec:resultsGENmove}

We finally test our SDP-based B\&B method on the ROPF problem with \eqref{GENmoves}. We compute several plans of active generation $P^0$ using random starting points. More precisely, we first compute a random value between $P_n^{min}$ and $P_n^{max}$ for each generator $n\in G$. Then we check if the sum of random active generations is greater or equal to 1.02 times the sum of real loads, that is, the total real load plus an estimation of the usual losses. If this constraint is satisfied, we stop. Otherwise, we proportionally reallocate the missing generation. However, due to the upper bound $P_n^{max}$, this redistribution may not be sufficient. In this case, we randomly go through the list of generators and increase those that still have margin until the constraint is satisfied. We propose to test 5 plans of generation for each instance. We focus on instances with more than 1000 buses. We only present instances for which we are certain that a feasible solution does exist.
To avoid the binary variables $\delta^-$ and $\delta^+$, we solve two problems for each instance: one with the variable $\lambda^-$ (i.e. $\delta^-=1$ and $\delta^+=0$) corresponding to a decrease in generation and the other with the variable $\lambda^+$ (i.e. $\delta^-=0$ and $\delta^+=1$) corresponding to an increase in generation.

Table \ref{ResultsGENmove1} presents the results for the MATPOWER instances with more than 1000 buses but less than 2850 buses. There are 7 instances and 5 active generation scenarios for each instance so 35 test cases in total.  The indication '+' or '-'  aims to specify if the best found solution corresponds to an increase or a decrease in generation. The rounding procedure solves 20 test cases out of 35. 11 test cases are solved within the first phase of our procedure and 9 more thanks to the B\&B procedure. Thus our procedure resolves as many instances as the rounding procedure. However, our procedure is more robust since it always finds a good feasible solution (with a gap less that 0.1\%) whereas there are 9 test cases for which the rounding procedure does not compute a feasible solution. It should be noted that our B\&B does not always improve the upper bound. This may be due to the time limit but since this is not an exact procedure, the time limit is a safeguard to avoid spending hours not improving anything.

Table \ref{ResultsGENmove2} presents the results for the MATPOWER instances with more than 2850 buses but less than 9000 buses. There are 8 instances and 5 active generation scenarios for each instance so 40 test cases in total. The columns are the same as in Table \ref{ResultsGENmove1}. For these instances, 8 test cases are solved with the rounding procedure against 6 with our procedure (2 within the first phase of the procedure and 4 with the B\&B procedure). Once again, our procedure is more robust since it finds a feasible solution for each test case whereas there are 21 instances with no solution with the rounding procedure. In particular, the rounding procedure is inefficient on the largest instances (with more than 6000 buses). On the contrary, the feasible solutions computed by our procedure are pretty good even for large instances: the optimality gap is always less than 0.1\%. However, the B\&B procedure does not improve the solution (or not significantly) for 34 out of 38 test cases, which also shows the limits of our method.

\section{Conclusion}
\label{sec:conclusion}
We have introduced three new variants of the ROPF problem that tackle RTE issues. We have proposed a SDP-based B\&B algorithm to solve them. This algorithm is very flexible since it is enough to adjust a few parameters to obtain a method adapted to each variant. Our algorithm is able to close the gap or at least improve it on most MATPOWER instances. Moreover, it often finds better solutions than the ones obtained with rounding heuristics and it is particularly robust on large instances (networks with more than 6000 buses). Indeed, rounding heuristics are often inefficient on large instances whereas our algorithm computes good feasible solutions in a reasonable amount of time. Our algorithm thus meets the needs of RTE which requires to compute good feasible solutions in a reasonable time and in a robust way. Future work consists in solving complexified versions of the ROPF problems presented in this paper, including other means of voltage control like transformer ratios or topological changes.


\begin{acknowledgements}
Many thanks to Rémy Clément, Stéphane Fliscounakis and Tanguy Janssen, research engineers at RTE, for the time they spent explaining RTE's issues to us.
\end{acknowledgements}

%
\section*{Funding}
This work was partially funded by the CIFRE grant n$^{\circ}$2018/0116.

\section*{Conflict of interest}
The authors declare that they have no conflict of interest.

\section*{Declarations}
MATPOWER instances are available on MATPOWER website \href{https://matpower.org/}{https://matpower.org/}. The ROPF datasets generated during and/or analysed during the current study are available from the corresponding author on request.

\bibliographystyle{spmpsci}      

\begin{thebibliography}{}
%
%
\bibitem{AMD}
Amestoy, P.R., Davis, T.A., Duff, I.S.: An approximate minimum degree ordering
  algorithm.
\newblock SIAM Journal on Matrix Analysis and Applications \textbf{17}(4),
  886--905 (1996)

\bibitem{Julia}
Bezanson, J., Edelman, A., Karpinski, S., Shah, V.B.: Julia: A fresh approach
  to numerical computing.
\newblock SIAM review \textbf{59}(1), 65--98 (2017)

\bibitem{binganeORPD}
Bingane, C., Anjos, M.F., Le~Digabel, S.: Tight-and-cheap conic relaxation for
  the optimal reactive power dispatch problem.
\newblock IEEE Transactions on Power Systems \textbf{34}(6), 4684--4693 (2019)

\bibitem{KNITRO}
Byrd, R.H., Nocedal, J., Waltz, R.A.: K nitro: An integrated package for
  nonlinear optimization.
\newblock In: Large-scale nonlinear optimization, pp. 35--59. Springer (2006)

\bibitem{capitanescu2010sensitivity}
Capitanescu, F., Wehenkel, L.: Sensitivity-based approaches for handling
  discrete variables in optimal power flow computations.
\newblock IEEE Transactions on Power Systems \textbf{25}(4), 1780--1789 (2010)

\bibitem{carpentier}
Carpentier, J.: Contribution to the economic dispatch problem.
\newblock Bulletin de la Societe Francaise des Electriciens \textbf{3}(8),
  431--447 (1962)

\bibitem{chen2017optimal}
Chen, G., Liu, L., Zhang, Z., Huang, S.: Optimal reactive power dispatch by
  improved gsa-based algorithm with the novel strategies to handle constraints.
\newblock Applied Soft Computing \textbf{50}, 58--70 (2017)

\bibitem{davoodi2019novel}
Davoodi, E., Babaei, E., Mohammadi-Ivatloo, B., Rasouli, M.: A novel fast
  semidefinite programming-based approach for optimal reactive power dispatch.
\newblock IEEE Transactions on Industrial Informatics \textbf{16}(1), 288--298
  (2019)

\bibitem{jump}
Dunning, I., Huchette, J., Lubin, M.: Jump: A modeling language for
  mathematical optimization.
\newblock SIAM Review \textbf{59}(2), 295--320 (2017)

\bibitem{fischetti2003local}
Fischetti, M., Lodi, A.: Local branching.
\newblock Mathematical programming \textbf{98}(1), 23--47 (2003)

\bibitem{AMPL}
Fourer, R., Gay, D.M., Kernighan, B.W.: A modeling language for mathematical
  programming.
\newblock Management Science \textbf{36}(5), 519--554 (1990)

\bibitem{fukudaI}
Fukuda, M., Kojima, M., Murota, K., Nakata, K.: Exploiting sparsity in
  semidefinite programming via matrix completion i: General framework.
\newblock SIAM Journal on Optimization \textbf{11}(3), 647--674 (2001).

\bibitem{heidari2017gaussian}
Heidari, A.A., Abbaspour, R.A., Jordehi, A.R.: Gaussian bare-bones water cycle
  algorithm for optimal reactive power dispatch in electrical power systems.
\newblock Applied Soft Computing \textbf{57}, 657--671 (2017)

\bibitem{hijazi2017convex}
Hijazi, H., Coffrin, C., Van~Hentenryck, P.: Convex quadratic relaxations for
  mixed-integer nonlinear programs in power systems.
\newblock Mathematical Programming Computation \textbf{9}(3), 321--367 (2017)

\bibitem{kayacik2020misocp}
Kayacik, S.E., Kocuk, B.: An misocp-based solution approach to the reactive
  optimal power flow problem.
\newblock IEEE Transactions on Power Systems  (2020)

\bibitem{kilicc2014optimizing}
K{\i}l{\i}{\c{c}}, U., Ayan, K., Arifo{\u{g}}lu, U.: Optimizing reactive power
  flow of hvdc systems using genetic algorithm.
\newblock International Journal of Electrical Power \& Energy Systems
  \textbf{55}, 1--12 (2014)

\bibitem{lenin2016hybrid}
Lenin, K., Reddy, B.R., Suryakalavathi, M.: Hybrid tabu search-simulated
  annealing method to solve optimal reactive power problem.
\newblock International Journal of Electrical Power \& Energy Systems
  \textbf{82}, 87--91 (2016)

\bibitem{lopez2021optimal}
Lopez, J.C., Rider, M.J., et~al.: Optimal reactive power dispatch with discrete
  controllers using a branch-and-bound algorithm: A semidefinite relaxation
  approach.
\newblock IEEE Transactions on Power Systems  (2021)

\bibitem{mccormick}
McCormick, G.P.: Computability of global solutions to factorable nonconvex
  programs: Part i—convex underestimating problems.
\newblock Mathematical programming \textbf{10}(1), 147--175 (1976)

\bibitem{ben2018whale}
ben~oualid Medani, K., Sayah, S., Bekrar, A.: Whale optimization algorithm
  based optimal reactive power dispatch: A case study of the algerian power
  system.
\newblock Electric Power Systems Research \textbf{163}, 696--705 (2018)

\bibitem{mehdinejad2016solution}
Mehdinejad, M., Mohammadi-Ivatloo, B., Dadashzadeh-Bonab, R., Zare, K.:
  Solution of optimal reactive power dispatch of power systems using hybrid
  particle swarm optimization and imperialist competitive algorithms.
\newblock International Journal of Electrical Power \& Energy Systems
  \textbf{83}, 104--116 (2016)

\bibitem{mosek}
Mosek: The mosek optimization software.
\newblock Online at http://www. mosek. com  (2019)

\bibitem{fukudaII}
Nakata, K., Fujisawa, K., Fukuda, M., Kojima, M., Murota, K.: Exploiting
  sparsity in semidefinite programming via matrix completion ii: Implementation
  and numerical results.
\newblock Mathematical Programming \textbf{95}(2), 303--327 (2003)

\bibitem{rajan2016exchange}
Rajan, A., Malakar, T.: Exchange market algorithm based optimum reactive power
  dispatch.
\newblock Applied Soft Computing \textbf{43}, 320--336 (2016)

\bibitem{OBBT}
Sahinidis, N.V.: Global optimization and constraint satisfaction: The
  branch-and-reduce approach.
\newblock In: International Workshop on Global Optimization and Constraint
  Satisfaction, pp. 1--16. Springer (2002)

\bibitem{survey_MINLP_nonconvex}
Samuel~Burer, A.L.L.: Non-convex mixed-integer nonlinear programming: A survey.
\newblock Surveys in Operations Research and Management Science \textbf{17},
  97--106 (2012).

\bibitem{sliwak2020clique}
{Sliwak}, J., {Andersen}, E.D., {Anjos}, M.F., {Létocart}, L., {Traversi}, E.:
  A clique merging algorithm to solve semidefinite relaxations of optimal power
  flow problems.
\newblock IEEE Transactions on Power Systems \textbf{36}(2), 1641--1644 (2021).

\bibitem{MathProgComplex_powertech}
{Sliwak}, J., {Ruiz}, M., {Anjos}, M.F., {Létocart}, L., {Traversi}, E.: A
  julia module for polynomial optimization with complex variables applied to
  optimal power flow.
\newblock In: 2019 IEEE Milan PowerTech, pp. 1--6 (2019).

\bibitem{chordal_graphs_SDP}
Vandenberghe, L., Andersen, M.S., et~al.: Chordal graphs and semidefinite
  optimization.
\newblock Foundations and Trends in Optimization \textbf{1}(4), 241--433 (2015)

\bibitem{xu2019optimal}
Xu, H., Dominguez-Garcia, A., Sauer, P.W.: Optimal tap setting of voltage
  regulation transformers using batch reinforcement learning.
\newblock IEEE Transactions on Power Systems  (2019)

\bibitem{yang2017optimal}
Yang, Z., Bose, A., Zhong, H., Zhang, N., Xia, Q., Kang, C.: Optimal reactive
  power dispatch with accurately modeled discrete control devices: A successive
  linear approximation approach.
\newblock IEEE Transactions on Power Systems \textbf{32}(3), 2435--2444 (2017)

\bibitem{zhou2014strength}
Zhou, B., Chan, K.W., Yu, T., Wei, H., Tang, J.: Strength pareto multigroup
  search optimizer for multiobjective optimal reactive power dispatch.
\newblock IEEE Transactions on Industrial Informatics \textbf{10}(2),
  1012--1022 (2014)

\bibitem{matpower}
Zimmerman, R.D., Murillo-S{\'a}nchez, C.E., Thomas, R.J., et~al.: Matpower:
  Steady-state operations, planning, and analysis tools for power systems
  research and education.
\newblock IEEE Transactions on power systems \textbf{26}(1), 12--19 (2011)
\end{thebibliography}


\pagestyle{empty}
\begin{landscape}
\begin{table}[htbp]
\caption{Results for the ROPF problem with constraint \eqref{MAXkshunts}}
\label{ResultsMAXkshunts}
\centering
\begin{tabular}{lrr|rrr|rrrrr|rr}
Instance  & $|S|$ & k & \multicolumn{3}{c|}{Computation of bounds} & \multicolumn{5}{c|}{B\&B} & \multicolumn{2}{c}{Rounding} \\
   &  &  & UB         & LB         & Gap     & \#binvar &  \#nodes & Time(s) & UB  & Gap & UB & Gap     \\
\hline
14            & 1        & 4    & 5371.50    & 5371.50    & \textbf{0.00\%}  &                   &               &               & & & 5371.50            & \textbf{0.00\%}  \\
24\_ieee\_rts & 1        & 4    & 44259.19   & 44259.19   & \textbf{0.00\%}  &                   &               &               & & & 44259.24           & \textbf{0.00\%}  \\
30            & 2        & 4    & 373.41     & 373.39     & \textbf{0.00\%}  &                   &               &               & & & 373.41             & \textbf{0.00\%}  \\
ieee30        & 2        & 4    & 5927.59    & 5927.59    & \textbf{0.00\%}  &                   &               &               & & & 5927.64            & \textbf{0.00\%}  \\
57            & 3        & 4    & 25337.79   & 25337.79   & \textbf{0.00\%}  &                   &               &                & & & 25337.70           & \textbf{0.00\%}  \\
89pegase      & 44       & 4    & 5813.41    & 5812.94    & \textbf{0.00\%} &                   &               &                & & & 5812.96            & \textbf{0.00\%}  \\
118           & 14       & 4    & 86301.50   & 86298.49   & \textbf{0.00\%}  &                   &               &              & &   & 86301.52           & \textbf{0.00\%}  \\
ACTIVSg200    & 4        & 4    & 34641.24   & 34640.46   & \textbf{0.00\%}  &                   &               &                & & & 34640.11           & \textbf{0.00\%}  \\
illinois200   & 4        & 4    & 43763.98   & 43763.92   & \textbf{0.00\%}  &                   &               &              & &   & 43763.98           & \textbf{0.00\%}  \\
300           & 29       & 4    & 503727.37  & 475470.69  & 5.61\%  & 7   &13&	32.51
              & 475482.55     & \textbf{0.00\%}         & 475526.23          & 0.01\%  \\
300mod        & 29       & 4    & 476470.28  & 475475.06  & 0.21\%  & 7    & 13	& 23.28
              & 475482.47     & \textbf{0.00\%}         & 503035.65          & 5.48\%  \\
ACTIVSg500    & 15       & 4    & 91446.78   & 90967.34   & 0.52\%  & 5  & 47&	102.17
                & 91446.78      & 0.52\%         & 91454.53           & 0.53\%  \\
1354pegase    & 1082     & 4    & Inf        & 74104.05   & -       & 5   & 1	& 18.80
               & 74107.29      & \textbf{0.00\%}         & 74114.91           & 0.01\%  \\
1888rte       & 45       & 4    & Inf        & 59621.43   & -       & 11  & 181 &	$>$3600
               & 59882.62      & 0.44\%         & 59986.33           & 0.61\%  \\               
1951rte       & 24       & 4    & 81838.43   & 81744.99   & 0.11\%  & 6  &5&	70.41
                & 81751.01      & \textbf{0.00\%}        & 81838.41           & 0.11\%  \\
ACTIVSg2000   & 149      & 4    & 1245341.82 & 1244050.72 & 0.10\%  & 7  & 23 &	$>$3600
               & 1245353.63   & 0.10\%         & Inf                & -       \\               
2736sp        & 1        & 4    & 1320569.85 & 1320559.84 & \textbf{0.00\%}  &                   &               &          & &       & 1320569.91         & \textbf{0.00\%}  \\
2737sop       & 5        & 4    & 793976.68  & 793733.53  & 0.03\%  & 5 & 5&	244.95
                 & 793749.57     & \textbf{0.00\%}         & 793749.57          & \textbf{0.00\%}  \\
2746wop       & 6        & 4    & 1223001.54 & 1222947.85 & \textbf{0.00\%}  &                   &               &        & &         & 1223001.58         & \textbf{0.00\%}  \\
2848rte       & 48       & 4    & 53052.04   & 53030.69   & 0.04\%  & 7   & 106 & $>$3600
               & 53051.60      & 0.04\%         & Inf                & -       \\               
2868rte       & 33       & 4    & 79836.41   & 79832.52   & \textbf{0.00\%}  &                   &               &        & &         & Inf                & -       \\
2869pegase    & 2197     & 12    & Inf        & 134155.54 & -    &  14 & 13 & 434.81 & 134155.82 &    \textbf{0.00\%}             & 134185.65        & 0.02\% \\
3012wp        & 9        & 4    & 2582246.71 & 2581933.31 & 0.01\%  & 5    &5&	457.80
              & 2582191.41    & \textbf{0.00\%}        & 2582191.41         &\textbf{0.00\%}  \\
3120sp        & 9        & 4    & 2141382.41 & 2139564.66 & 0.08\%  & 7 & 38 &	$>$3600      & 2141382.41   & 0.08\%         & Inf                & -       \\                 
3375wp        & 9        & 4    & 7404662.46 & 7404199.17 & \textbf{0.00\%} &                   &               &           & &      & 7404269.18         & \textbf{0.00\%}  \\
6468rte       & 97       & 4    & 86938.67   & 86869.63   & 0.08\%  & 8&23&	$>$3600
                 & 86938.67      & 0.08\%         & Inf                & -       \\                 
6470rte       & 73       & 4    & Inf        & 98650.43   & -       & 6 & 23&	$>$3600
                & Inf      & -         & Inf                & -       \\                
6495rte       & 99       & 14   & 106740.51  & 106310.40  & 0.40\%  & 17 & 26 &	$>$3600
               & 106740.51     & 0.40\%         & Inf                & -       \\               
6515rte       & 102      & 66   & 110238.10  & 109913.67  & 0.29\%  & 74 & 26	&$>$3600
                & 110238.10     & 0.29\%         & Inf                & -       \\                
9241pegase    & 7327     & 125 & 315980.89  & 311438.46  & 1.44\%  & 165 & 18 &$>$3600               &    315980.89        &   1.44\%            & Inf                & -       \\
13659pegase   & 8754     & 1100 & 385680.71  & 380743.60  & 1.28\%  & 1237 & 28 & $>$3600              & 385680.71             & 1.28\%              & 385675.83          & 1.28\% 
\end{tabular}
\end{table} 
\end{landscape}
\pagestyle{plain}

\pagestyle{empty}
\begin{landscape}
\begin{table}[htbp]
\caption{Results for the ROPF problem with constraint \eqref{MAXkmoves} for $k=4$}
\label{ResultsMAXkmoves}
\centering
\begin{tabular}{ll|rrr|rrrrr|rr}
Instance  & $|S|$ & \multicolumn{3}{c|}{Computation of bounds} & \multicolumn{5}{c|}{B\&B} & \multicolumn{2}{c}{Rounding} \\
   &  & UB         & LB         & Gap                            & \#binvar                 & \#nodes                  & Time (s)                 & UB   & Gap                      &             UB              & Gap                            \\
\hline
14            & 1        & 5371.50    & 5371.50    & \textbf{0.00\%} &  &  &  &            &                                            & 5371.50                    & \textbf{0.00\%} \\
24\_ieee\_rts & 1        & 44259.19   & 44259.19   & \textbf{0.00\%} &  &  &  &            &                                            & 44259.24                   & \textbf{0.00\%} \\
30            & 2        & 373.41     & 373.39     & \textbf{0.00\%} &  &  &  &            &                                            & 373.41                     & \textbf{0.00\%} \\
\_ieee30      & 2        & 5927.59    & 5927.59    & \textbf{0.00\%} &  &  &  &            &                                            & 5927.64                    & \textbf{0.00\%} \\
57            & 3        & 25337.79   & 25337.79   & \textbf{0.00\%} &  &  &  &            &                                            & 25337.70                   & \textbf{0.00\%} \\
89pegase      & 44       & 5812.69    & 5812.69    & \textbf{0.00\%} &  &  &  &            &                                            & 5812.69                    & \textbf{0.00\%} \\
118           & 14       & 86298.82   & 86296.28   & \textbf{0.00\%} &  &  &  &            &                                            & 86299.64                   & \textbf{0.00\%} \\
\_ACTIVSg200  & 4        & 34641.15   & 34640.46   & \textbf{0.00\%} &  &  &  &            &                                            & 34640.11                   & \textbf{0.00\%} \\
\_illinois200 & 4        & 43763.98   & 43763.92   & \textbf{0.00\%} &  &  &  &            &                                           & 43763.98                   & \textbf{0.00\%} \\
300           & 29       & 475394.34  & 475379.12  & \textbf{0.00\%} &  &  &  &            &                                        & 475395.09                  & \textbf{0.00\%} \\
300mod        & 29       & 475396.68  & 475382.09  & \textbf{0.00\%} &  &  &  &            &                                         & 475395.68                  & \textbf{0.00\%} \\
\_ACTIVSg500  & 15       & 91439.37   & 90782.00   & 0.72\%                         &         0                 &      1                    &    17.38                     &    91343.17        &  0.61\%                               & 91343.17                   & 0.61\%                         \\
1354pegase    & 1082     & 74049.71   & 74045.79   & \textbf{0.00\%} &  &  &  &            &                                &             74049.81                   & \textbf{0.00\%} \\
1888rte       & 45       & 59840.90   & 59610.98   & 0.38\%                         & 10                       & 336                      & \textgreater{}3600      & 59825.10 & 0.36\%                         &             59841.60                   & 0.39\%                         \\
1951rte       & 24       & 81746.93   & 81736.90   & 0.01\%                         & 5                        & 7                        & 70.36                    & 81742.62   & \textbf{0.00\%} &             81748.30                   & 0.01\%                         \\
\_ACTIVSg2000 & 149      & 1242319.11 & 1241812.60 & 0.04\%                         & 54                       & 17                       & 1512.15                  & 1242134.91 & 0.03\%                         &             1242202.90                 & 0.03\%                         \\
2736sp        & 1        & 1320569.85 & 1320553.46 & \textbf{0.00\%} &  &  &  &            &                                &             1320569.91                 & \textbf{0.00\%} \\
2737sop       & 5        & 793976.68  & 793735.65  & 0.03\%                         & 1                        & 1                        & 57.35                    & 793749.61  & \textbf{0.00\%}                         &             793746.84                  & \textbf{0.00\%} \\
2746wop       & 6        & 1223001.54 & 1222994.89 & \textbf{0.00\%} &  &  &  &            &                                &             1223001.58                 & \textbf{0.00\%} \\
2848rte       & 48       & 53043.10   & 53027.08   & 0.03\%                         & 27                       & 62                      & \textgreater{}3600      & 53042.10   & 0.03\%                         &             Infeasible & -                              \\
2868rte       & 33       & 79829.59   & 79827.34   & \textbf{0.00\%} &  &  &  &            &                                &             Inf                        & -                              \\
2869pegase    & 2197     & 133953.39  & 133948.52  & \textbf{0.00\%} &  &  &  &            &                                &            133953.35                  & \textbf{0.00\%} \\
3012wp        & 9        & 2582185.13 & 2581915.43 & \textbf{0.00\%} &  &  &  &            &                                &            Infeasible & -                              \\
3120sp        & 9        & Inf        & 2139522.22 & -                              & 2                        & 5                        & 829.46                   & 2141455.77 & 0.09\%                                     & Inf                        & -                              \\
3375wp        & 9        & 7404267.03 & 7404180.26 & \textbf{0.00\%} &  &  &  &            &                                           & Inf                        & -                              \\
6468rte       & 97       & Inf        & 86863.18   & -                              & 41                       & 20                       & \textgreater{}3600      & 86889.13   & 0.03\%                                     & Inf                        & -                              \\
6470rte       & 73       & Inf        & 98633.73   & -                              & 26                       & 14                       & 3417.24                  & 98642.4    & \textbf{0.00\%} &             Inf                        & -                              \\
6495rte       & 99       & Inf        & 106303.77  & -                              & 42                       & 15                       & \textgreater{}3600      & 106724.29  & 0.39\%                                     & Inf                        & -                              \\
6515rte       & 102      & 110253.47  & 109921.01  & 0.30\%                         & 30                       & 18                      & \textgreater{}3600     & 110250.40   & 0.30\%                        & Inf                        & -                              \\
9241pegase    & 7327     & 315674.22  & 311071.20  & 1.46\%                         & 1549                     & 12                        & \textgreater{}3600                         & 315674.22  & 1.46\%                          & Inf                        & -                              \\
13659pegase   & 8754     & Inf        & 380668.34  & -                              & 2283                     & 10                       & \textgreater{}3600      & 385572.62   & 1.27\%                                    & Inf                        & - 
\end{tabular}
\end{table}
\end{landscape}
\pagestyle{plain}

\pagestyle{empty}
\begin{landscape}
\begin{table}[htbp]
\caption{Results for the ROPF problem with constraint \eqref{GENmoves} for instances with less than 2850 buses}
\label{ResultsGENmove1}
\centering
\begin{tabular}{lll|rrr|rrrrr|rr}
Instance  & $|S|$ &  +/- & \multicolumn{3}{c|}{Computation of bounds} & \multicolumn{5}{c|}{B\&B} & \multicolumn{2}{c}{Rounding} \\
& & & UB & LB  & Gap  & \#binvar & \#nodes & Time (s) & UB  & Gap & UB  & Gap\\ 
\hline
1354pegase & 1082  & - & 74454.57 & 74446.65 & 0.01\% & 301 &  & \textgreater{}3600 & 74454.57 & 0.01\% & 74450.93 & \textbf{0.00\%} \\
1354pegase & 1082  & + & 74560.06 & 74550.55 & 0.01\% & 270 & 871 & \textgreater{}3600 & 74560.06 & 0.01\% & 74556.06 & \textbf{0.00\%} \\
1354pegase & 1082  & + & 74668.53 & 74654.90 & 0.02\% & 283 & 883 & \textgreater{}3600 & 74665.98 & 0.01\% & 74661.81 & \textbf{0.00\%} \\
1354pegase & 1082  & - & 74594.60 & 74583.01 & 0.02\% & 329 & 805 & \textgreater{}3600 & 74594.60 & 0.02\% & 74587.74 & \textbf{0.00\%} \\
1354pegase & 1082  & + & 74603.54 & 74592.73 & 0.01\% & 327 & 906 & \textgreater{}3600 & 74603.54 & 0.01\% & 74597.49 & \textbf{0.00\%} \\
\hline
1951rte & 24 & - & Inf & 82091.31 & - & 5 & 21 & 296.10 & 82095.94 & \textbf{0.00\%} & Inf & - \\
1951rte & 24 & - & 82204.12 & 82194.85 & 0.01\% & 5 & 7 & 84.77 & 82198.36 & \textbf{0.00\%} & Inf & - \\
1951rte & 24 & - & 82234.35 & 82229.82 & \textbf{0.00\%} &  &  &  &  &  & 82240.59 & 0.01\% \\
1951rte & 24 & - & Inf & 82060.86 & - & 7 & 19 & 289.74 & 82066.48 & \textbf{0.00\%} & Inf & - \\
1951rte & 24 & - & Inf & 82176.29 & - & 5 & 24 & 409.15 & 82183.64 & \textbf{0.00\%} & Inf & - \\
\hline
ACTIVSg2000 & 149  & + & 68554.86 & 68514.50 & 0.06\% & 109 & 53 & \textgreater{}3600 & 68554.86 & 0.06\% & 68554.38 & 0.06\% \\
ACTIVSg2000 & 149  & + & 68525.15 & 68489.58 & 0.05\% & 115 & 47 & \textgreater{}3600 & 68525.15 & 0.05\% & 68525.69 & 0.05\% \\
ACTIVSg2000 & 149  & - & 68443.64 & 68407.81 & 0.05\% & 114 & 52 & \textgreater{}3600 & 68443.64 & 0.05\% & 68444.34 & 0.05\% \\
ACTIVSg2000 & 149  & - & 68438.09 & 68400.88 & 0.05\% & 113 & 54 & \textgreater{}3600 & 68438.09 & 0.05\% & 68438.90 & 0.06\% \\
ACTIVSg2000 & 149  & + & 68463.58 & 68451.39 & 0.02\% & 112 & 52 & \textgreater{}3600 & 68463.58 & 0.02\% & 68465.25 & 0.02\% \\
\hline
2736sp & 1  & - & 18363.68 & 18363.44 & \textbf{0.00\%} &  &  &  &  &  & 18363.68 & \textbf{0.00\%} \\
2736sp & 1  & - & 18369.37 & 18369.26 & \textbf{0.00\%} &  &  &  &  &  & 18369.37 & \textbf{0.00\%} \\
2736sp & 1  & - & 18378.56 & 18378.36 & \textbf{0.00\%} &  &  &  &  &  & 18378.56 & \textbf{0.00\%} \\
2736sp & 1  & - & 18366.73 & 18366.45 & \textbf{0.00\%} &  &  &  &  &  & 18366.73 & \textbf{0.00\%} \\
2736sp & 1  & - & 18371.55 & 18371.36 & \textbf{0.00\%} &  &  &  &  &  & 18371.55 & \textbf{0.00\%} \\
\hline
2737sop & 5  & - & 11418.83 & 11416.91 & 0.02\% & 1 & 2 & 78.76 & 11417.15 & \textbf{0.00\%} & 11417.12 & \textbf{0.00\%} \\
2737sop & 5  & - & 11419.29 & 11416.84 & 0.02\% & 1 & 2 & 64.42 & 11417.07 & \textbf{0.00\%} & 11416.99 & \textbf{0.00\%} \\
2737sop & 5  & - & 11413.85 & 11411.50 & 0.02\% & 1 & 2 & 62.63 & 11411.74 & \textbf{0.00\%} & 11411.67 & \textbf{0.00\%} \\
2737sop & 5  & - & 11417.89 & 11415.66 & 0.02\% & 1 & 2 & 70.24 & 11415.92 & \textbf{0.00\%} & 11415.84 & \textbf{0.00\%} \\
2737sop & 5  & - & 11416.65 & 11414.10 & 0.02\% & 1 & 2 & 79.26 & 11414.38 & \textbf{0.00\%} & 11414.30 & \textbf{0.00\%} \\
\hline
2746wop & 6  & - & 19273.40 & 19273.16 & \textbf{0.00\%} &  &  &  &  &  & 19273.40 & \textbf{0.00\%} \\
2746wop & 6  & - & 19263.41 & 19263.15 & \textbf{0.00\%} &  &  &  &  &  & 19263.41 & \textbf{0.00\%} \\
2746wop & 6  & - & 19289.75 & 19289.39 & \textbf{0.00\%} &  &  &  &  &  & 19289.59 & \textbf{0.00\%} \\
2746wop & 6  & - & 19286.57 & 19286.04 & \textbf{0.00\%} &  &  &  &  &  & 19286.57 & \textbf{0.00\%} \\
2746wop & 6  & - & 19285.37 & 19284.97 & \textbf{0.00\%} &  &  &  &  &  & 19285.25 & \textbf{0.00\%} \\
\hline
2848rte & 48  & - & Inf & 53142.85 & - & 34 & 140 & \textgreater{}3600 & 53163.39 & 0.04\% & Inf & - \\
2848rte & 48  & - & Inf & 53185.72 & - & 31 & 130 & \textgreater{}3600 & 53206.27 & 0.04\% & Inf & - \\
2848rte & 48  & - & Inf & 53170.47 & - & 33 & 132 & \textgreater{}3600 & 53193.82 & 0.04\% & Inf & - \\
2848rte & 48  & - & Inf & 53191.46 &  & 29 & 149 & \textgreater{}3600 & 53210.87 & 0.04\% & Inf & - \\
2848rte & 48  & - & 53201.04 & 53175.97 & 0.05\% & 30 & 135 & \textgreater{}3600 & 53201.04 & 0.05\% & Inf & - \\
\end{tabular}
\end{table}
\end{landscape}
\pagestyle{plain}

\pagestyle{empty}
\begin{landscape}
\begin{table}[htbp]
\caption{Results for the ROPF problem with constraint \eqref{GENmoves} for instances with more than 2850 buses}
\label{ResultsGENmove2}
\centering
\begin{tabular}{lll|rrr|rrrrr|rr}
Instance  & $|S|$  & +/- & \multicolumn{3}{c|}{Computation of bounds} & \multicolumn{5}{c|}{B\&B} & \multicolumn{2}{c}{Rounding} \\
& & & UB & LB  & Gap  & \#binvar & \#nodes & Time (s) & UB  & Gap & UB  & Gap\\ 
\hline
2868rte & 33  & - & 80021.98 & 80016.75 & \textbf{0.00\%} &  &  &  &  &  & Inf & - \\
2868rte & 33  & - & Inf & 80100.29 & - & 8 & 10 & 261.76 & 80101.64 & \textbf{0.00\%} & 80101.64 & \textbf{0.00\%} \\
2868rte & 33  & - & Inf & 80113.82 & - & 9 & 11 & 2652.89 & 80116.18 & \textbf{0.00\%} & 80115.97 & \textbf{0.00\%} \\
2868rte & 33  & - & Inf & 80098.29 & - & 9 & 11 & 256.93 & 80100.42 & \textbf{0.00\%} & 80100.42 & \textbf{0.00\%} \\
2868rte & 33  & - & Inf & 80155.48 & - & 9 & 28 & 735.94 & 80158.09 & \textbf{0.00\%} & Inf & - \\
\hline
2869pegase & 2197  & - & 134844.35 & 134829.63 & 0.01\% & 577 & 307 & \textgreater{}3600 & 134844.35 & 0.01\% & 134836.89 & \textbf{0.00\%} \\
2869pegase & 2197  & + & 135468.07 & 135451.33 & 0.01\% & 438 & 345 & \textgreater{}3600 & 135468.0721 & 0.01\% & 135459.81 & \textbf{0.00\%} \\
2869pegase & 2197  & + & 135221.95 & 135193.34 & 0.02\% & 471 & 170 & \textgreater{}3600 & 135221.95 & 0.02\% & 135202.13 & \textbf{0.00\%} \\
2869pegase & 2197  & - & 134999.53 & 134976.67 & 0.02\% & 493 & 308 & \textgreater{}3600 & 134999.53 & 0.02\% & 134985.34 & \textbf{0.00\%} \\
2869pegase & 2197  & + & 135378.32 & 135357.60 & 0.02\% & 511 & 345 & \textgreater{}3600 & 135378.32 & 0.02\% & 135366.56 & \textbf{0.00\%} \\
\hline
3012wp & 9  & + & 27771.36 & 27755.06 & 0.06\% & 2 & 7 & 285.85 & 27770.66 & 0.06\% & 27770.66 & 0.06\% \\
3012wp & 9  & + & 27844.23 & 27831.10 & 0.05\% & 1 & 3 & 109.16 & 27844.10 & 0.05\% & 27844.10 & 0.05\% \\
3012wp & 9  & + & 27810.15 & 27795.77 & 0.05\% & 1 & 3 & 116.11 & 27809.83 & 0.05\% & 27809.83 & 0.05\% \\
3012wp & 9  & + & 27767.51 & 27752.24 & 0.05\% & 1 & 3 & 156.41 & 27767.26 & 0.05\% & 27767.28 & 0.05\% \\
3012wp & 9  & + & 27799.43 & 27784.16 & 0.05\% & 1 & 3 & 125.86 & 27798.93 & 0.05\% & 27798.93 & 0.05\% \\
\hline
3120sp & 9   & + & 21642.54 & 21627.63 & 0.07\% & 2 & 7 & 326.77 & 21642.43 & 0.07\% & 21642.43 & 0.07\% \\
3120sp & 9   & + & 21779.92 & 21765.93 & 0.06\% & 1 & 3 & 124.55 & 21779.87 & 0.06\% & 21779.87 & 0.06\% \\
3120sp & 9   & + & 21691.65 & 21677.24 & 0.07\% & 2 & 7 & 308.49 & 21691.65 & 0.07\% & 21691.67 & 0.07\% \\
3120sp & 9   & + & 21678.85 & 21663.55 & 0.07\% & 2 & 7 & 258.10 & 21678.85 & 0.07\% & 21678.93 & 0.07\% \\
3120sp & 9   & + & 21635.82 & 21620.20 & 0.07\% & 2 & 5 & 238.77 & 21635.24 & 0.07\% & 21635.24 & 0.07\% \\
\hline
6468rte & 97  & + & 87158.54 & 87105.29 & 0.06\% & 64 & 55 & \textgreater{}3600 & 87158.54 & 0.06\% & Inf & - \\
6468rte & 97  & + & 87044.03 & 87022.41 & 0.02\% & 71 & 48 & \textgreater{}3600 & 87044.03 & 0.02\% & Inf & - \\
6468rte & 97  & + & 87138.68 & 87086.74 & 0.06\% & 68 & 51 & \textgreater{}3600 & 87138.68 & 0.06\% & Inf & - \\
6468rte & 97  & + & 87215.35 & 87215.35 & \textbf{0.00\%} &  &  &  &  &  & Inf & - \\
6468rte & 97  & + & 87285.51 & 87246.14 & 0.05\% & 65 & 45 & \textgreater{}3600 & 87285.51 & 0.05\% & Inf & - \\
\hline
6470rte & 73 & + & 98745.81 & 98733.04 & 0.01\% & 36 & 50 & \textgreater{}3600 & 98745.81 & 0.01\% & Inf & - \\
6470rte & 73  & + & 98831.51 & 98796.99 & 0.03\% & 47 & 50 & \textgreater{}3600 & 98831.51 & 0.03\% & Inf & - \\
6470rte & 73  & + & 98706.23 & 98683.05 & 0.02\% & 47 & 51 & \textgreater{}3600 & 98706.23454 & 0.02\% & Inf & - \\
6470rte & 73  & + & 98698.39 & 98677.03 & 0.02\% & 43 & 48 & \textgreater{}3600 & 98698.39249 & 0.02\% & Inf & - \\
6470rte & 73  & + & 98758.59 & 98748.08 & 0.01\% & 42 & 50 & \textgreater{}3600 & 98758.59 & 0.01\% & Inf & - \\
\hline
6495rte & 99 & - & 106406.16 & 106383.20 & 0.02\% & 47 & 30 & \textgreater{}3600 & 106406.16 & 0.02\% & Inf & - \\
6495rte & 99  & + & 106318.61 & 106302.86 & 0.01\% & 44 & 37 & \textgreater{}3600 & 106318.61 & 0.01\% & Inf & - \\
6495rte & 99  & + & 106585.90 & 106573.25 & 0.01\% & 40 & 52 & \textgreater{}3600 & 106585.9033 & 0.01\% & Inf & - \\
6495rte & 99  & + & 106519.60 & 106506.92 & 0.01\% & 46 & 36 & \textgreater{}3600 & 106519.5953 & 0.01\% & Inf & - \\
6495rte & 99  & + & 106556.86 & 106535.65 & 0.02\% & 43 & 36 & \textgreater{}3600 & 106556.86 & 0.02\% & 106551.00 & 0.01\% \\
\hline
6515rte & 102  & + & 110106.56 & 110084.21 & 0.02\% & 33 & 35 & \textgreater{}3600 & 110106.56   & 0.02\% & Inf & - \\
6515rte & 102  & + & 110034.86 & 110008.20 & 0.02\% & 32 & 37 & \textgreater{}3600 & 110034.86   & 0.02\% & Inf & - \\
6515rte & 102  & + & 110066.07 & 110045.04 & 0.02\% & 32 & 40 & \textgreater{}3600 & 110066.07   & 0.02\% & Inf & - \\
6515rte & 102   & + & 110114.50 & 110092.21 & 0.02\% & 33 & 42 & \textgreater{}3600 & 110114.5045 & 0.02\% & Inf & - \\
6515rte & 102   & + & 110128.23 & 110102.90 & 0.02\% & 31 & 43 & \textgreater{}3600 & 110128.23   & 0.02\% & Inf & -
\end{tabular}
\end{table}
\end{landscape}
\pagestyle{plain}

\end{document}